\documentclass[final]{cvpr}
\usepackage{amsmath}
\usepackage{amssymb}
\usepackage{times}
\usepackage{bm}
\usepackage{xargs}
\usepackage{multirow}
\usepackage{amsfonts}
\usepackage{color}
\usepackage{booktabs}
\newcommandx\includeImageLineWidth[2][1=1.0]{\includegraphics[width=#1\linewidth]{#2}}

\usepackage{array}
\newcommand{\PreserveBackslash}[1]{\let\temp=\\#1\let\\=\temp}
\newcolumntype{C}[1]{>{\PreserveBackslash\centering}p{#1}}
\newcolumntype{R}[1]{>{\PreserveBackslash\raggedleft}p{#1}}
\newcolumntype{L}[1]{>{\PreserveBackslash\raggedright}p{#1}}

\usepackage{times}
\usepackage{epsfig}
\usepackage{graphicx}
\usepackage{amsmath}
\usepackage{amssymb}
\usepackage{subfiles}
\usepackage{subfigure}
\usepackage{booktabs} 
\usepackage{amsfonts}   
\usepackage{algorithm}
\usepackage{algorithmic}
\usepackage{enumitem}
\usepackage{soul}
\newcommand\Mark[1]{\textsuperscript#1}

\usepackage[pagebackref=true,breaklinks=true,colorlinks,bookmarks=false]{hyperref}

\definecolor{burntorange}{rgb}{0.8, 0.33, 0.0}
\definecolor{amber}{rgb}{1.0, 0.75, 0.0}
\def\cvprPaperID{10384} % *** Enter the CVPR Paper ID here
\def\confYear{CVPR 2021}

\begin{document}

%%%%%%%%% TITLE
\title{Multimodal Contrastive Training for Visual Representation Learning}

\author{Xin Yuan\Mark{1}\thanks{This work has been done during the first author’s internship at Adobe.}, Zhe Lin\Mark{2}, Jason Kuen\Mark{2}, Jianming Zhang\Mark{2}, Yilin Wang\Mark{2}\\
Michael Maire\Mark{1}, Ajinkya Kale\Mark{2}, and Baldo Faieta\Mark{2}\\
\Mark{1}University of Chicago \Mark{2}Adobe Research \\
{\tt\small \{yuanx, mmaire\}@uchicago.edu \{zlin, kuen, jianmzha, yilwang, akale, bfaieta\}@adobe.com}
}

\maketitle

\maketitle

\begin{abstract}
We develop an approach to learning visual representations that embraces multimodal data, driven by a combination of intra- and inter-modal similarity preservation objectives. Unlike existing visual pre-training methods, which solve a proxy prediction task in a single domain, our method exploits intrinsic data properties within each modality and semantic information from cross-modal correlation simultaneously, hence improving the quality of learned visual representations. By including multimodal training in a unified framework with different types of contrastive losses, our method can learn more powerful and generic visual features. We first train our model on COCO and evaluate the learned visual representations on various downstream tasks including image classification, object detection, and instance segmentation.  For example, the visual representations pre-trained on COCO by our method achieve state-of-the-art top-1 validation accuracy of $55.3\%$ on ImageNet classification, under the common transfer protocol. We also evaluate our method on the large-scale Stock images dataset and show its effectiveness on multi-label image tagging, and cross-modal retrieval tasks.
%For example, the visual representations pre-trained on COCO by our method achieve state-of-the-art top-1 validation accuracy of $55.3\%$ on ImageNet classification, under the common transfer protocol. Experiments across ImageNet, PASCAL VOC, COCO, and Stock datasets for image classification, tagging, object detection, instance segmentation, and cross-modal retrieval demonstrate that we learn high-quality visual features with better scalability and transferability.
%\zl{Maybe we can make the last two sentences more precise....like this: For experiments, we first train our model on COCO and demonstrate the superior performance of our learned representations on a number of downstream tasks including image classification, object detection, and semantic segmentation. For example, the visual representations pre-trained on COCO by our method achieve state-of-the-art top-1 validation accuracy of $55.3\%$ on ImageNet classification, under the common transfer protocol. We also evaluate our method on large-scale Stock images dataset and show its effectiveness on multi-label image tagging, and text-to-image retrieval tasks}

\end{abstract}

\section{Introduction}
\begin{figure}[t]
   \begin{minipage}[b]{\linewidth}
   \subfigure[Sup.]{
      \includegraphics[height=0.33\columnwidth]{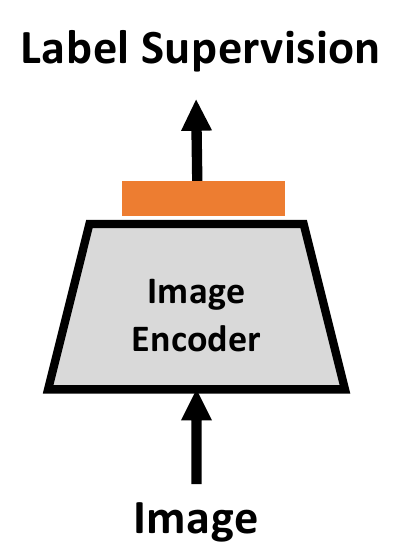}
   }
   \hfill
   \subfigure[Self-Sup]{
      \includegraphics[height=0.33\columnwidth]{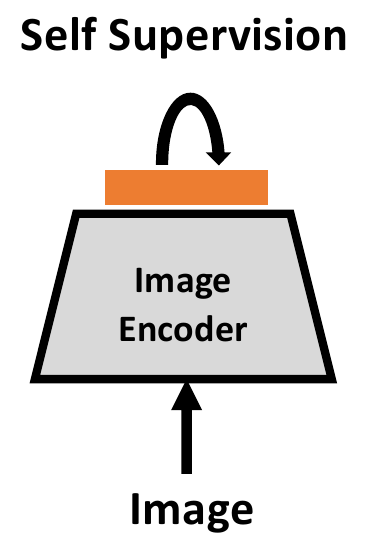}
   }
   \hfill
   \subfigure[VirTex~\cite{DBLP:journals/corr/abs-2006-06666}]{
      \includegraphics[height=0.33\columnwidth]{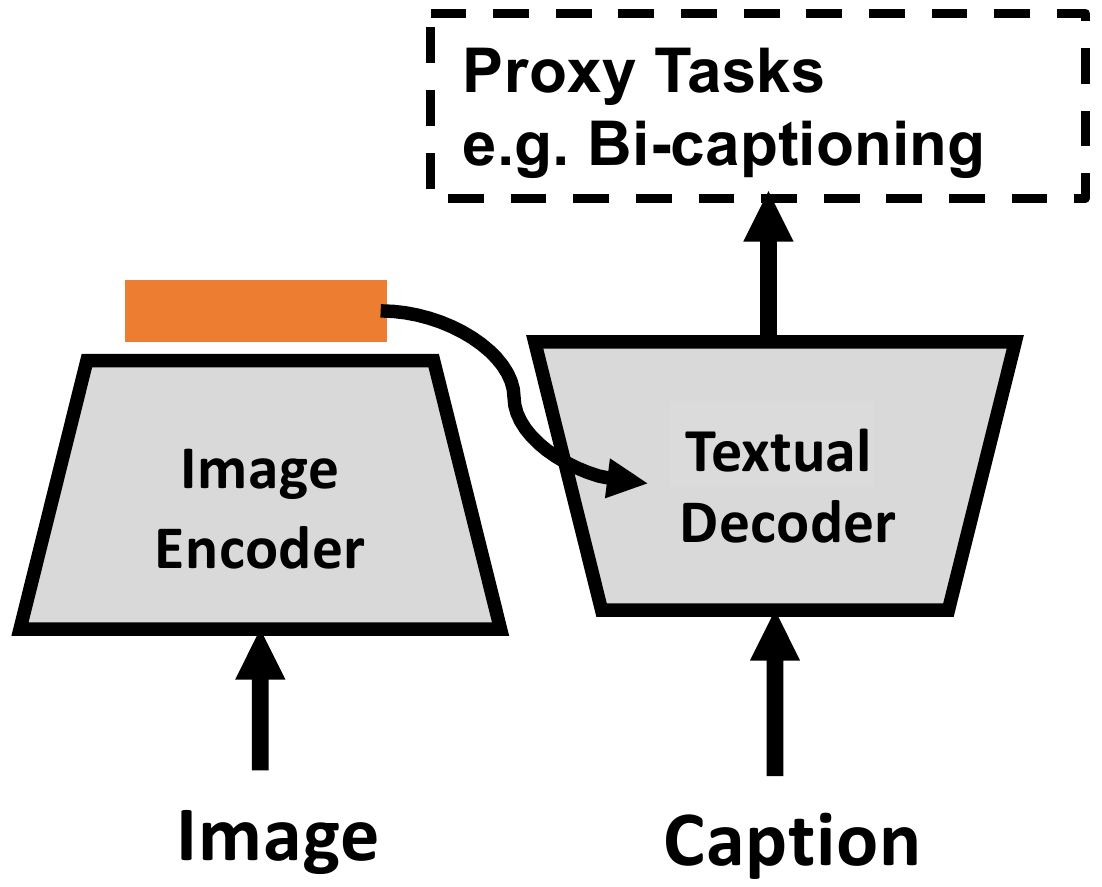}
   }
  \hfill
  \begin{center}
   \subfigure[Our Method]{
      \includegraphics[height=0.33\columnwidth]{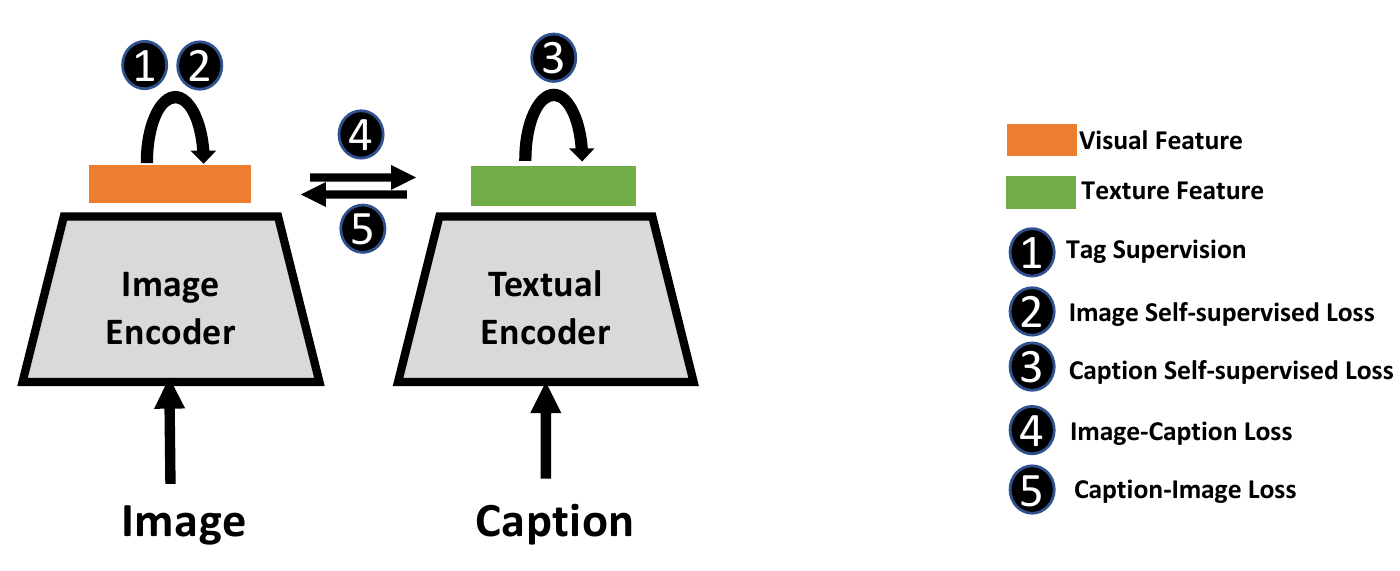}
   }
   \end{center}
   \caption{
     Main idea of the proposed method. Different from (c) VirTex~\cite{DBLP:journals/corr/abs-2006-06666}, our method not only learns the cross-modal correlation between images and captions, but also exploits intrinsic data properties in a self-supervised manner within each modality. 
   }
   \label{fig:motivation}
   \end{minipage}
\vspace{-2em}
\end{figure}
Visual representation learning is crucial for many computer vision tasks including image classification~\cite{DBLP:conf/cvpr/DengDSLL009,DBLP:journals/corr/SimonyanZ14a,DBLP:conf/cvpr/HeZRS16,DBLP:conf/cvpr/HuangLMW17}, tagging~\cite{DBLP:conf/cvpr/GattupalliZL19,DBLP:conf/cvpr/0002ZFY019}, object detection~\cite{DBLP:conf/cvpr/GirshickDDM14,DBLP:conf/cvpr/RedmonDGF16,DBLP:conf/eccv/LiuAESRFB16}, semantic and instance segmentation~\cite{DBLP:conf/cvpr/LongSD15,DBLP:conf/iccv/HeGDG17}.
Supervised pre-training over large-scale datasets~\cite{DBLP:conf/cvpr/DengDSLL009} yields useful visual features which lead to state-of-the-art performance on those tasks.
Yet, fine-grained class labeling efforts~\cite{DBLP:conf/cvpr/DengDSLL009} are prohibitively heavy. Self-supervised learning methods~\cite{DBLP:conf/iccv/CaronBMJ19,DBLP:conf/iccv/DoerschGE15,DBLP:conf/eccv/ZhangIE16,DBLP:conf/cvpr/He0WXG20,DBLP:journals/corr/abs-2002-05709,DBLP:journals/corr/abs-2003-04297} do not require any annotations, but still require either extremely large training sets or longer training epochs.
%to close the gaps between unsupervised and supervised representation learning.

In addition to labels, image data usually comes with additional information including tags and captions, which is typically generated by internet users and therefore easier to acquire.
More importantly, such multimodal information comes with higher-level abstract concepts, which offer the potential for drawing useful connections across different modalities~\cite{DBLP:conf/iccv/GuWCC17,DBLP:conf/cvpr/KarpathyL15,DBLP:journals/corr/KirosSZ14,DBLP:conf/cvpr/GuCJN018, DBLP:conf/bmvc/FaghriFKF18}.

Our objective is to learn visual representation from multimodal data in a unified training framework.
The framework design should have the following essential properties: (1) fully exploits data potential within each unlabeled modality in a self-supervised manner; (2) bridges the heterogeneity gap by comparing different modalities in a common semantic space with similarity preservation objectives; (3) can be easily extended to take any new incoming modality.
We aim to learn high-quality visual features, which benefit from not only the additional semantic information learned by cross-modal correlation modeling, but also the intrinsic data properties provided by each modality itself. 

Some recently proposed methods~\cite{DBLP:conf/cvpr/QuattoniCD07,DBLP:conf/iccv/LiJJM17,DBLP:conf/cvpr/Gomez-BigordaPR17,DBLP:conf/cvpr/GordoL17,DBLP:journals/corr/abs-2006-06666, DBLP:conf/eccv/SariyildizPL20,DBLP:conf/nips/Alwassel0KTGT20} also focus on generating high-quality visual representations from scratch using multimodal data.
For example, VirTex~\cite{DBLP:journals/corr/abs-2006-06666} makes a trade-off between the data efficiency and annotation effort by relaxing the extremeness of the unsupervised setting and embracing caption annotations which are relatively easy to acquire.
However, as shown in Figure~\ref{fig:motivation}, VirTex is still trained in a single-path manner by solving a cross-modal proxy task, which is not sufficient to exploit the full potential within each individual modality.

\begin{figure*}[tb]
% \vskip 0.2in
\begin{center}
\centerline{\includegraphics[width=1.99\columnwidth]{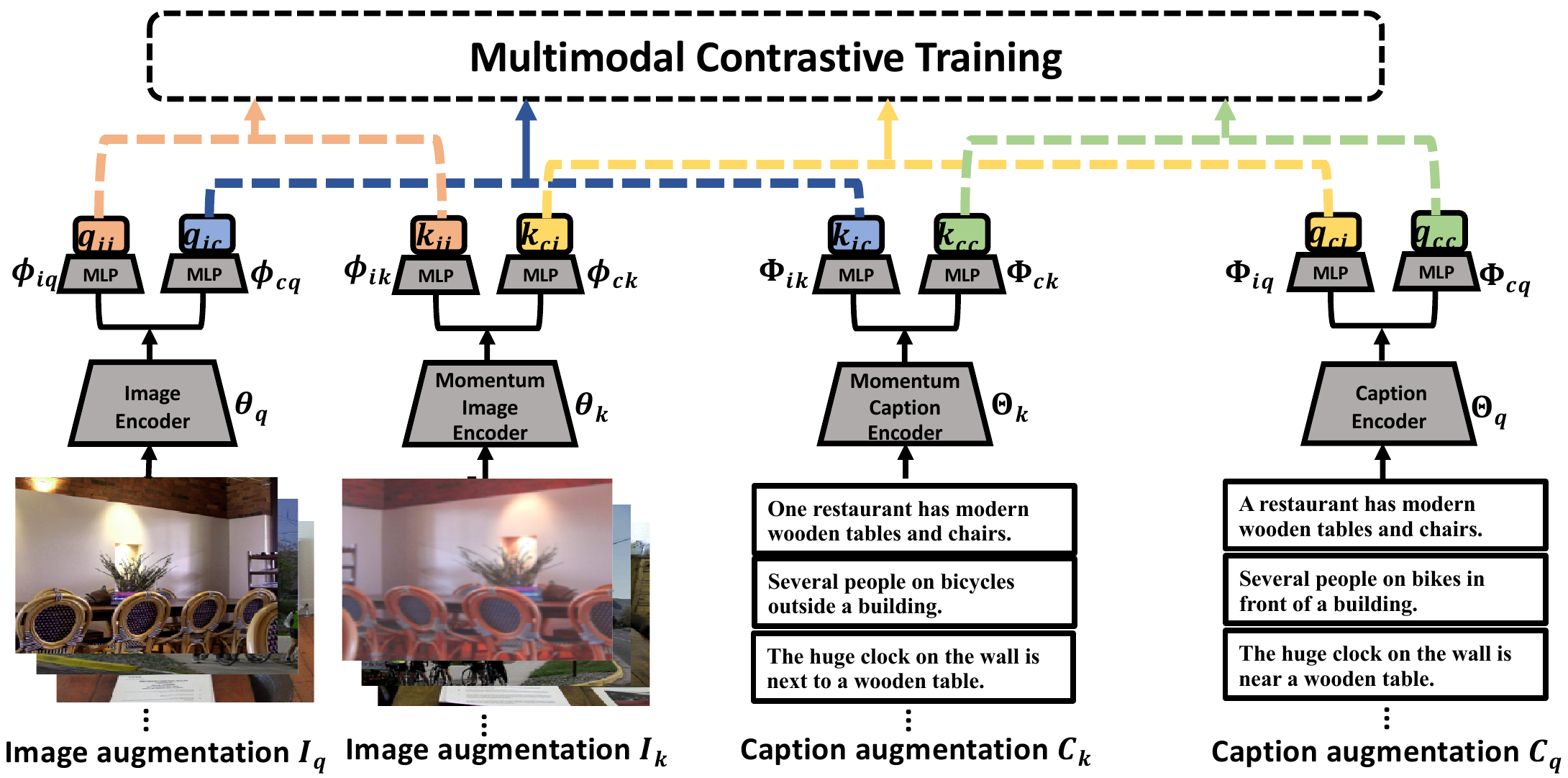}}
\caption{Framework of our proposed method is composed of two contrastive training schemes: intra-modal(%contrastive learning (the 
\textcolor{burntorange}{\textbf{orange}} and \textcolor{green}{\textbf{green}} paths) and inter-modal (%the 
\textcolor{amber}{\textbf{yellow}} and \textcolor{blue}{\textbf{blue}} paths) contrastive learning. 
In intra-modal contrastive learning, we train encoders for each individual modality in a self-supervised manner. In inter-modal contrastive learning, we compare different modalities in a common embedding space with bi-directional similarity preservation objectives. \textit{Best viewed in color.}}
\label{fig:framework}
\end{center}
\vspace{-3em}
\end{figure*}

In this paper, we take a unified view of both intra- and inter-modal similarity preservation in multi-modal data, based on which we develop a new visual representation learning framework, as shown in Figure~\ref{fig:motivation}.
To be specific, an intra-modal training path is used to capture the intrinsic patterns of augmented data examples in a prediction task.
A inter-modal training scheme is used to enhance the visual features by embracing the cross-modal interactions.
With carefully designed contrastive losses, features in all modalities are adjusted via backpropagation in multiple training paths.
We summarize our contributions as two-fold:
\begin{itemize}[leftmargin=.15in]
   \item{%
      \textbf{Unified Multi-modal Training.}~
      Our multi-modal training framework can exploit intrinsic data properties within each modality and extract semantic information from cross-modal correlation simultaneously. In addition, as shown in Figure~\ref{fig:motivation} and~\ref{fig:framework}, our framework is symmetric for all modalities, which suggests it has the flexibility to incorporate any new modality. 
   }%
   \item{%
      \textbf{Broad Transferability.}~
      The visual representations pre-trained by our method can be transferred to many downstream computer vision tasks and achieve excellent performance under the common transfer learning protocols. 
      %Note that the main focus of this paper is visual representation learning, even though our method also generate useful features in other modality (\textit{e.g.} texture features) as a by-product.
   }%
\end{itemize}
We demonstrate these advantages through making experimental comparisons between supervised, self-supervised and learning from text methods through extensive experiments on classification, tagging, cross-modal retrieval, object detection, and instance segmentation.

\section{Related Work}
\noindent \textbf{Self-supervised learning.}
%Noise contrastive estimation~\cite{} is commonly used in~\cite{} instances which is a special form of contrastive learning.
% 2 61 46 36 66 35 56
Many self-supervised methods~\cite{DBLP:conf/nips/BachmanHB19,DBLP:journals/corr/abs-1807-03748,DBLP:conf/iclr/HjelmFLGBTB19,DBLP:conf/iccv/ZhuangZY19,DBLP:journals/corr/abs-1905-09272,DBLP:conf/cvpr/He0WXG20,DBLP:journals/corr/abs-2002-05709,DBLP:journals/corr/abs-2003-04297, DBLP:journals/corr/abs-1906-05849, DBLP:conf/cvpr/WuXYL18} utilize contrastive objectives for instance comparison in order to facilitate visual representation learning. 
For example, ~\cite{DBLP:conf/cvpr/WuXYL18} use a memory bank which stores previously-computed representations and the noise-contrastive estimation (NCE)~\cite{DBLP:journals/jmlr/GutmannH10} to tackle the computational challenges imposed by the large number of instance classes.
MoCo~\cite{DBLP:conf/cvpr/He0WXG20} further improve such a scheme by storing representations from a momentum encoder in dynamic dictionary with a queue.
SimCLR~\cite{DBLP:journals/corr/abs-2002-05709} propose a simple framework under the large-batch setting, removing the needs of memory representations. 
MoCov2~\cite{DBLP:journals/corr/abs-2003-04297} borrow the multi-layer perceptron (MLP) head design from~\cite{DBLP:journals/corr/abs-2002-05709} and show  significant improvements.
Our method shares the same spirit with these methods, in that we both use contrastive visual representation learning. However, we embrace multimodal data in multiple training paths to better align the visual features with additional semantic information. \\
\textbf{Joint visual-textual pretraining.}
Vision and language (VL) methods~\cite{DBLP:conf/iccv/GuWCC17,DBLP:conf/cvpr/KarpathyL15,DBLP:journals/corr/KirosSZ14,DBLP:conf/cvpr/GuCJN018,DBLP:conf/cvpr/WangLL16,DBLP:conf/nips/LuBPL19,DBLP:conf/emnlp/TanB19,DBLP:conf/eccv/ChenLYK0G0020,li2019visualbert,gu2020self} are representatives that embrace multi-modal information for many computer vision tasks, such as image captioning and cross-modal retrieval. 
Such methods aim at mapping text and images into a common space, where semantic similarity across different modalities can be learned by ranking-based contrastive losses.
Many works~\cite{DBLP:conf/iclr/SuZCLLWD20,DBLP:conf/eccv/Li0LZHZWH0WCG20} also focus on learning a joint visual-textual space via a fine-tuned BERT for VL tasks.
However, these methods depend on the pre-trained image feature extractors or object detectors instead of learning from scratch on target datasets.
Recently, ~\cite{DBLP:journals/corr/abs-2006-16228} utilizes the NCE~\cite{DBLP:journals/jmlr/GutmannH10} and MIL-NCE losses~\cite{DBLP:conf/cvpr/MiechASLSZ20} to learn representations using across video, language and audio modalities. 
Such methods share the similar cross-modal similarity preservation concept with ours in loss design. However, they only consider cross-modal correlation mining, while ours focuses on both intra-modal and inter-modal learning.
Additionally, we demonstrate data efficiency by using a smaller dataset for pre-training visual representations.
\\
\textbf{Language-guided visual representation learning.} 
Recently, several works~\cite{DBLP:conf/cvpr/QuattoniCD07,DBLP:conf/iccv/LiJJM17,DBLP:conf/cvpr/Gomez-BigordaPR17,DBLP:conf/cvpr/GordoL17,DBLP:journals/corr/abs-2006-06666, DBLP:conf/eccv/SariyildizPL20} propose pre-training approaches that use semantically dense captions to learn visual representations from scratch.
For example, Virtex~\cite{DBLP:journals/corr/abs-2006-06666} jointly trains a convolutional network and a Transformer~\cite{DBLP:conf/nips/VaswaniSPUJGKP17} from scratch to generate natural language captions for images, \textit{i.e.} formulating a bi-directional image captioning proxy task.
ICMLM~\cite{DBLP:conf/eccv/SariyildizPL20} introduces image-conditioned masked language modeling (ICMLM) as a proxy task to learn visual representations over image-caption pairs. Both methods yield visual representations with good scalability and transferablity.
However, the proxy tasks in these methods work in a single-path manner by merely conditioning on the visual representation while ignoring the intrinsic properties of visual data itself.
\\

\section{Method}
\begin{figure*}[tb]
   \begin{minipage}[b]{\linewidth}
   \subfigure[Visual(Textual) Contrastive Loss]{
      \includegraphics[height=0.3\columnwidth]{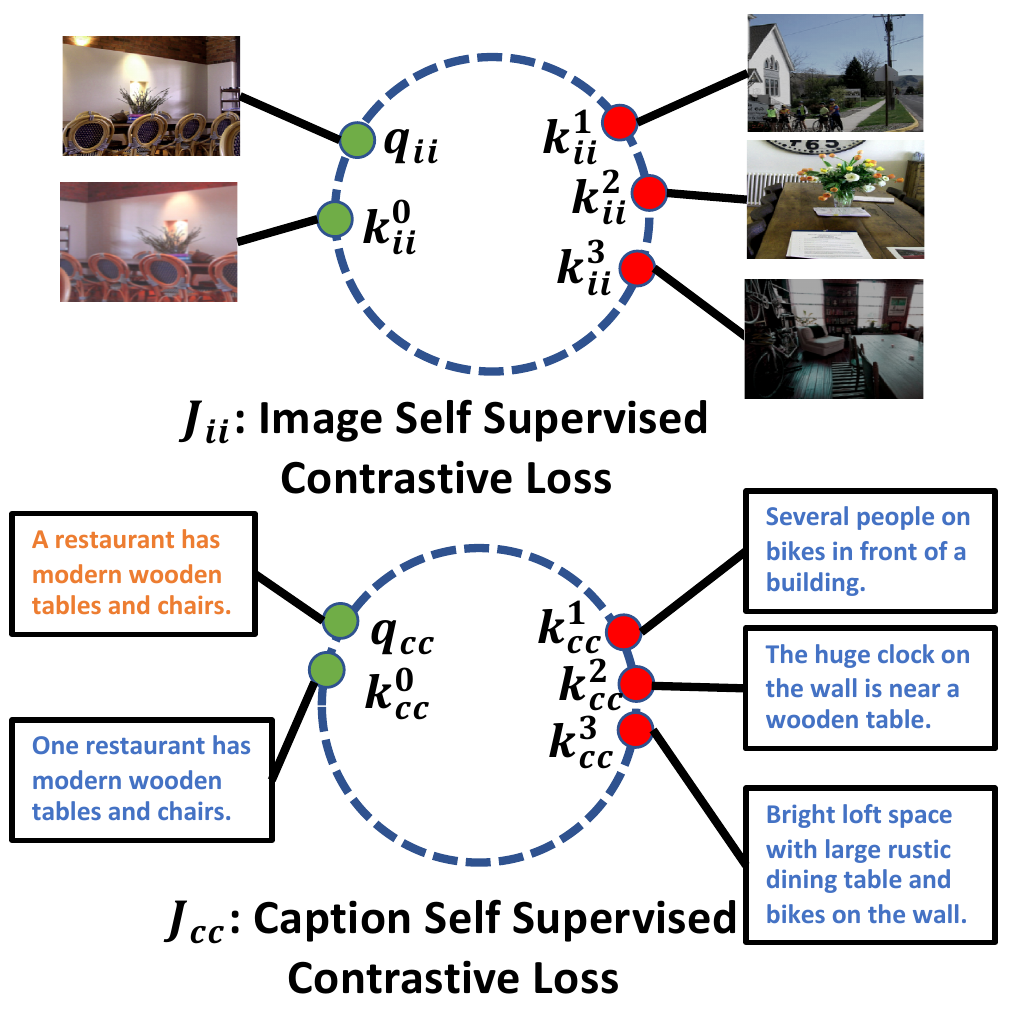}
   }
   \hfill
   \subfigure[Tag-supervised Contrastive Loss]{
      \includegraphics[height=0.3\columnwidth]{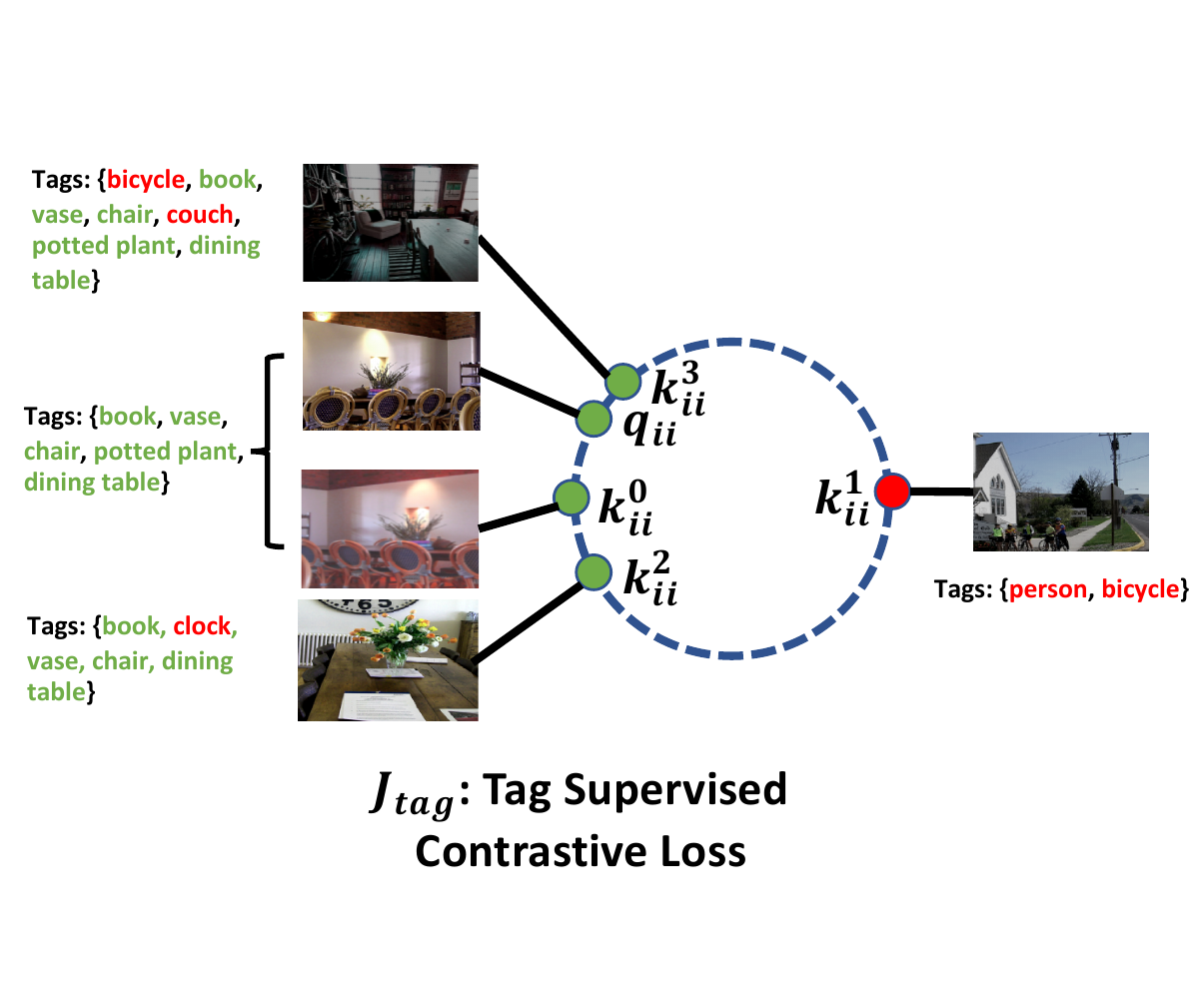}
   }
   \hfill
   \subfigure[Visual-Semantic Contrastive Loss]{
      \includegraphics[height=0.3\columnwidth]{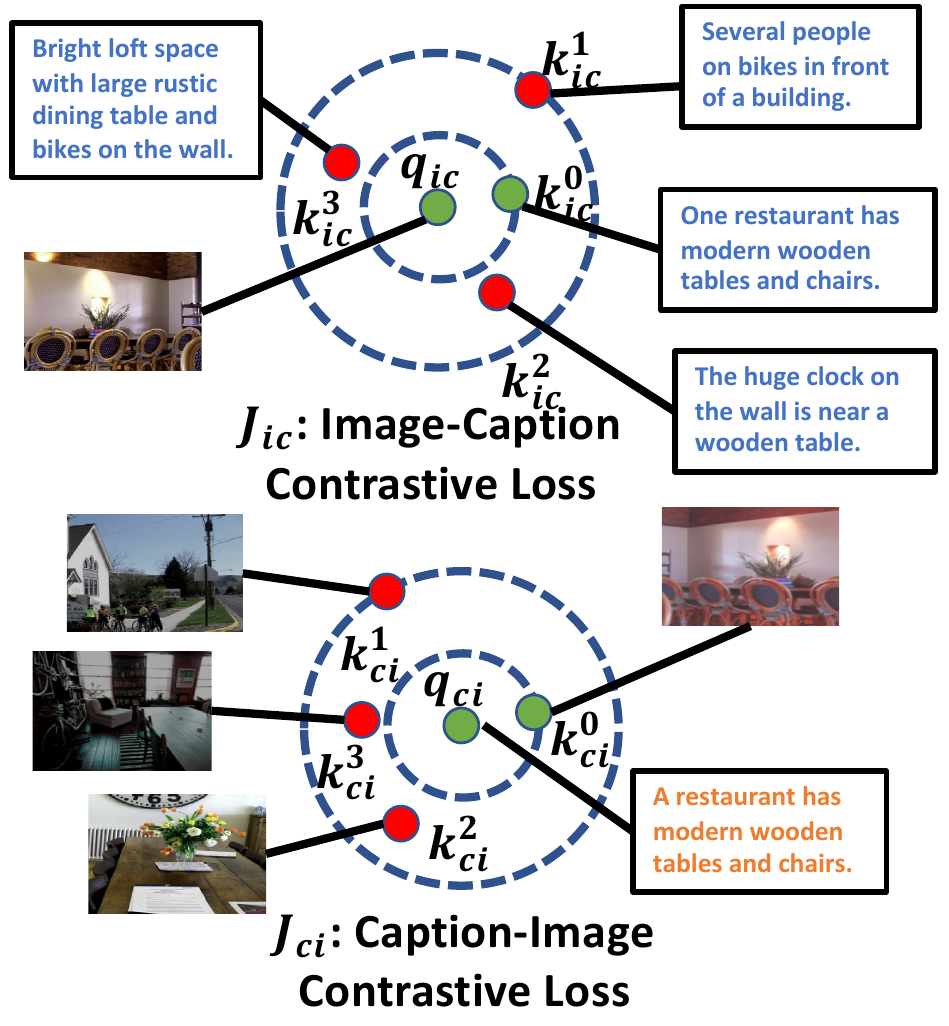}
   }
   \caption{
     Different types of contrastive loss in our method. These losses enable similarity preservation in both intra- and inter-modal training, encouraging features of semantically-similar input examples to be closer. 
   }
   \label{fig:losses}
   \end{minipage}
\vspace{-2em}
\end{figure*}

%\subsection{System Overview}
Figure~\ref{fig:framework} shows the overall architecture for the proposed multi-modal contrastive training framework. 
The system is composed of two contrastive training schemes: 
intra-modal (orange and green paths) and intra-modal (yellow and blue paths) contrastive learning with different types of losses which are shown in Figure~\ref{fig:losses}.
The intra-modal training scheme is based on existing self-supervised representation learning framework MoCo-v2\cite{DBLP:journals/corr/abs-2003-04297} that captures the intrinsic patterns of augmented image examples.
However, self-supervised methods lack of the ability to learn semantic information from higher-level concepts~\cite{DBLP:journals/corr/abs-2004-11362}.
We address such limitation by (1) designing additional textual encoder and momentum encoder to capture semantic information from augmented sentences. (2) involving the tag information in the contrastive loss to improve the visual representations.

The inter-modal training scheme is designed to enhance the visual feature by embracing the cross-modal interactions. 
We first embed the visual and textual features into a common space. 
Then, we design a visual-semantic contrastive loss to force the features of semantically-similar input examples to be closer. 
As such, visual features will be adjusted according to the captions via back propagation, and vice versa.
Note that we use distinct MLP layers for cross-modal feature embedding so that the two intra-modality and inter-modality training schemes do not interfere with each other.
Through the combinations of these two training schemes, 
we can learn powerful and generic visual representations. 
Although the proposed method also generates useful textual features as a by-product, it is not the main focus of this paper.
After the multi-modal contrastive training has completed, the visual encoder can be directly applied to, or fine-tuned for, various downstream tasks.
\subsection{Intra-modality Contrastive Learning}
We first denote the multi-modal dataset as $D=\{(I_j,c_j, t_j)\}$, which %is composed of
comprises $N$ image-caption-tags tuples. 
Note that $t_j$ is a $K$-dim binary vector where $t_j^{(k)}$ is an indicator of the occurrence of a specific $k$-th tag in $I_j$. 
Our intra-modality contrastive training aims to preserve the similarity within the augmented variants of the same image or caption through self-supervised learning. As a running example, we formulate intra-modal visual/textual contrastive learning based on the MoCo-v2 framework. \\
\noindent \textbf{Visual Contrastive Learning.}
We parameterize the image encoder as $f_{iq}(\cdot;\theta_q,\phi_{iq})$ and momentum image encoder as $f_{ik}(\cdot,\theta_k, \phi_{ik})$, where $\theta$ and $\phi$ are the weights of convolutional neural network (CNN) backbone and 2-layer MLP head, respectively.
% embed different augmentations of the image examples to -dim \textit{backbone} features. 
The weights $\theta_k, \phi_{ik}$ are updated with momentum coefficient $m$: $\theta_k \leftarrow m\theta_k + (1-m)\theta_q, \phi_{ik} \leftarrow m\phi_{ik} + (1-m)\phi_{iq}$. 
The notation differs from MoCo-v2, which takes the encoder weights as a whole, since we require to map backbone features into different spaces, decoupling the feature embeddings from intra-modal and inter-modal training paths.
For augmented examples $I_j^{\dagger}, I_j^{\star}$ from the same input image  $I_j$ in a minibatch, image encoder and momentum encoder embed them to \textit{query} and \textit{key} features:
\begin{eqnarray} \label{feat_ii}
q_{ii}^{j} = f_{iq}(I_j^{\dagger};\theta_q,\phi_{iq}) \label{feat_qii} \\
k_{ii}^{j} = f_{ik}(I_j^{\star};\theta_k,\phi_{ik}) \label{feat_kii}
\end{eqnarray}
Then, a dynamic set of \textit{key} features with length $K$ %can be
is maintained by iterative \textit{dequeue} and \textit{enqueue} operations.
For \textit{query} feature $q_{ii}$ in the current mini-batch, $key$ feature $k_{ii}$ in the queue is denoted as $k_{ii}^{+}$ if it can form a positive pair with $q_{ii}$, \textit{i.e.} they originate from the same image.
The visual self-supervised contrastive loss shown in the top of Figure~\ref{fig:losses}(a) is defined as:
\begin{eqnarray}\label{loss_ii}
J_{ii} = -log \frac{exp(q_{ii}\cdot k_{ii}^{+}/\tau)}{\sum_{j=0}^{K}exp(q_{ii}\cdot k_{ii}^{j}/\tau)}
\end{eqnarray}
where $\cdot$ computes similarity scores between example pairs and $\tau$ is  temperature hyperparameter.

Self-supervised learning frameworks consider all augmented examples originated from %different sources
other images as negative samples, even if two images share very similar semantic concepts (e.g. they have a big overlap of tags)~\cite{DBLP:journals/corr/abs-2004-11362}. To encourage more closely semantic-aligned visual representation learning, 
we design an additional loss term in the visual-contrastive training path using the tag annotations provided by the dataset, as shown in Figure~\ref{fig:losses}(b).
For a \textit{query} image $I_j$, in addition to the sample originating from the same input source, we also consider $I_p$ as positive samples if $I_p$ shares some common tags with $I_j$. 
Formally, we extend the \textit{key} set from $\{k_{ii}^{+}\}$ to:
\begin{eqnarray}
P = \{k_{ii}^p  ~|~ \forall p: t_p\cdot t_j > \epsilon \}
\end{eqnarray}
where the dot product computes the similarity score between two tag lists and $\epsilon$ is a threshold hyperparameter.
We thus define the tag supervised loss term by modifying Eq.~\ref{loss_ii}:
\begin{eqnarray}
J_{tag} = -\frac{1}{|P|}\sum_{p \in P}log \frac{exp(q_{ii}\cdot k_{ii}^p/\tau)}{\sum_{j=0}^{K}exp(q_{ii}\cdot k_{ii}^{j}/\tau)}
\end{eqnarray}
where $|P|$ denotes the set size of P. Note that $J_{tag}$ degenerates to $J_{ii}$ when there are no samples in \textit{queue} sharing common semantics with \textit{query} sample, \textit{i.e.} $P=\{k_{ii}^{+}\}, |P|=1$. \\
\noindent\textbf{Textual Contrastive Learning.}
To learn useful semantic information from higher-level concepts, 
we design the textual encoder and momentum encoder to extract features from augmented captions.
For the textual encoder architecture, we use BERT-like transformer architecture~\cite{DBLP:conf/naacl/DevlinCLT19} with a 2-layer MLP head as a running example.
Formally, we parameterize the textual encoder as $f_{cq}(\cdot;\Theta_q,\Phi_{cq})$ and momentum textual encoder as $f_{ck}(\cdot,\Theta_k, \Phi_{ik})$, where $\Theta$ and $\Phi$ is the weights of transformer and 2-layer MLP head, respectively.
We utilize back-translation~\cite{DBLP:conf/emnlp/EdunovOAG18} for caption data augmentation.
As in visual contrastive learning, we also maintain the same notion of \textit{key}, \textit{query} and \textit{queue} in the textual contrastive training scheme.
Given different augmented examples $c_j^{\dagger}, c_j^{\star}$ from the same input caption source $c_j$ in a minibatch, the textural encoder and momentum encoder respectively embed them to \textit{query} and \textit{key} features. We formulate the embedding and mapping of caption modality as:
\begin{eqnarray} \label{feat_cc}
q_{cc}^{j} = f_{cq}(c_j^{\dagger};\Theta_q,\Phi_{cq}) \label{feat_qcc}  \\
k_{cc}^{j} = f_{ck}(c_j^{\star};\Theta_k,\Phi_{ck}) \label{feat_kcc} 
\end{eqnarray}
As shown in bottom of Figure~\ref{fig:losses}(a), we aim to predict the positive \textit{key} feature $k_{cc}^{+}$ from the \textit{queue} which originates from the same input source with $q_{cc}$.
The contrastive loss is thus defined as:  
\begin{eqnarray}
J_{cc} = -log \frac{exp(q_{cc}\cdot k_{cc}^{+}/\tau)}{\sum_{j=0}^{K}exp(q_{cc}\cdot k_{cc}^{j}/\tau)}
\end{eqnarray}
where the dot product denotes similarity score and $\tau$ is a temperature parameter.

\subsection{Inter-modality Contrastive Learning}
To utilize the semantic information from captions for better visual feature learning, we enable cross-modal interactions via an inter-modality contrastive training scheme.
We first embed the representations of the image and the caption into a common space and then use a ranking-based contrastive loss~\cite{DBLP:conf/cvpr/GuCJN018, DBLP:conf/bmvc/FaghriFKF18} to learn both the visual and textual model parameters. 
In particular, we use CNN and BERT-like transformer as representation model backbones, with additional distinct branches of MLP layers with %higher
larger output dimensions.
Note that such distinct MLPs design is not \textit{ad-hoc}, but based on the observation during experiments that using either unified or separate MLPs with the same-sized embedding space %will
degrades the downstream task performance.
As shown in Figure~\ref{fig:losses}(c), %we hope 
the objective functions %to 
encourage the %similarity
similarities of ground-truth caption-image pairs to be greater than %that 
those of all other negative pairs, instead of merely solving a \textit{hard} prediction task. \\
%As such, we consider the pairwise ranking-based contrastive loss which is composed of image-caption and caption-image contrastive loss.
\noindent\textbf{Image-to-Caption Contrastive Learning.}
Given an image-caption pair $(I_j, c_j)$, we generate the \textit{query} feature using image encoder and \textit{key} feature using momentum textual encoder, then map them to the common space:
\begin{eqnarray} \label{feat_ic}
q_{ic}^{j} = f_{iq}(I_j^{\dagger};\theta_q,\phi_{cq})\\
k_{ic}^{j} = f_{ck}(c_j^{\star};\Theta_k,\Phi_{ik})
\end{eqnarray}
where $\phi_{cq}, \Phi_{ik}$ denote distinct MLP layers parameters from $\phi_{iq}, \Phi_{ck}$ in Eq.~\ref{feat_qii} and Eq.~\ref{feat_kcc}.
We denote the positive \textit{key} feature $k_{ic}^{+}$ from the \textit{queue} which originates from the positive image-caption pair with $q_{ic}$, \textit{i.e.} image is described by the caption. In the common space, we aim to simultaneously minimize the distance between $q_{ic}$ and $k_{ic}^{+}$ and maximize the distances between $q_{ic}$ and all other negative \textit{key} features from the \textit{queue}. We thus formulate the image-caption contrastive loss as:
\begin{eqnarray}
J_{ic} = \sum_{j=1}^{K} [\alpha - q_{ic}\cdot k_{ic}^{+} + q_{ic}\cdot k_{ic}^j]_{+}
\end{eqnarray}
where $\alpha$ is the margin, the dot product denotes similarity score, and $[x]_{+}$ represents $max(x,0)$. \\
\noindent\textbf{Caption-to-Image Contrastive Learning.}
Similar with image-to-caption contrastive learning, we generate the \textit{query} feature using textual encoder and \textit{key} feature using momentum image encoder as:
\begin{eqnarray} \label{feat_ic}
q_{ci}^{j} = f_{cq}(c_j^{\dagger};\Theta_q,\Phi_{iq}) \\
k_{ci}^{j} = f_{ik}(I_j^{\star};\theta_k,\phi_{ck})
\end{eqnarray}
where $\Phi_{iq}, \phi_{ck}$ denote distinct MLP layers parameters from $\Phi_{cq}, \phi_{ik}$ in Eq.~\ref{feat_qcc} and Eq.~\ref{feat_kii}.
The caption-to-image contrastive loss which aims at optimizing the distance between caption \textit{query} and image \textit{queue} is defined as: 
\begin{eqnarray}
J_{ci} = \sum_{j=1}^{K} [\alpha - q_{ci}\cdot k_{ci}^{+} + q_{ci}\cdot k_{ci}^j]_{+}
\end{eqnarray}
in which $\alpha$ is the margin, the dot product is similarity score, and $[x]_{+}$ represents $max(x,0)$.

To this end, we formulate the final loss for our multimodal contrastive training method as:
\begin{eqnarray}
J = \lambda_{ii}J_{ii} + \lambda_{tag}J_{tag} + \lambda_{cc}J_{cc} + \lambda_{ic}J_{ic} + \lambda_{ci}J_{ci}   
\end{eqnarray}
where $\lambda_{ii}$, $\lambda_{tag}$,$\lambda_{cc}$, $\lambda_{ic}$, and $\lambda_{ci}$ are trade-off parameters among different contrastive losses.
Note that our method doesn't require all images to have tags. For some images with only tags or captions, the feature learning is guided by $\lambda_{tag}J_{tag} + \lambda_{ii}J_{ii}$ or $\lambda_{ii}J_{ii} + \lambda_{ci}J_{ci}+\lambda_{ic}J_{ic}+\lambda_{cc}J_{cc}$.
% \subsection{Joint Optimization}

% \begin{algorithm}[tb]
%   \caption{: Multi-modal Contrastive Training}
%   \label{alg:example}
% \begin{algorithmic}
%   \STATE {\bfseries Input:}
% \end{algorithmic}
% \end{algorithm}

\section{Experiments}
We evaluate performance on multiple downstream tasks.
\subsection{Experimental Setup}
\noindent \textbf{Pretraining Datasets.} 
We train our models on the image-caption-tag tuples of the 2017 split of COCO~\cite{DBLP:conf/eccv/LinMBHPRDZ14} and Stock~\cite{DBLP:journals/ijcv/ZhangMSSBLSPM17} image datasets. 
COCO  has  123K  images  (118K  and  5K  for  training  and  validation, respectively) with 5 captions for each image.  For COCO dataset, we create tag sets for each image using the object list from the instance annotations. (80 objects in total).
Stock is a large-scale dataset with 5.8 million training images and 50K test images.
For each image, we have a title and a list of tags associated with it.  We choose 18157 most frequent tags as our vocabulary during training.

\noindent\textbf{Implementation Details.}
For image modality, we use ResNet-50~\cite{DBLP:conf/cvpr/HeZRS16} as the backbone. We apply average pooling at the last layer of the backbone to generate a 2048-d feature vector.
We follow the data augmentation scheme in~\cite{DBLP:journals/corr/abs-2003-04297} to generate a 224 $\times$224 image crop by random resizing, color jittering, horizontal flipping, grayscale conversion and gaussian blurring. For caption modality, we use BertTokenizer and $\text{Bert}_{base}$ model~\cite{DBLP:conf/naacl/DevlinCLT19} to generate the 768-d pooled output, which is further processed by 2-layer MLP heads.
For the text branch, we use back-translation~\cite{DBLP:conf/emnlp/EdunovOAG18} for caption data augmentation. 
More specifically, the given English sentence is randomly translated to French or German then back to English during training using the machine translation models~\cite{DBLP:conf/wmt/OttEGA18, DBLP:conf/emnlp/EdunovOAG18}.
For all encoders, we use two distinct 2-layer MLP heads (hidden layer is 2048-d, with ReLU) to generate the last representations for contrastive training. 
The last layer of each encoder consists of a 128-d intra-modal representation and a 1024-d  inter-modal representation.
We normalize all feature vectors before calculating their dot products in the contrastive losses, where $\tau$ is set as 0.07. 
Momentum encoders' updating parameter $m$ is set as 0.999 while $K$ is 32768 and 65536 for COCO and Stock, respectively.
For trade-off parameters in the final loss, we set $\lambda_{ii}$ as 1.0, $\lambda_{ic}$ as $1e^{-4}$, $\lambda_{ci}$ as $1e^{-4}$ and $\lambda_{cc}$ as 1.0. For $\lambda_{tag}$, we set it to 1.0 for our method with tag otherwise set it to 0.
For loss term $J_{tag}$, we choose the threshold $\epsilon$ as 2 to extend the positive sample sets according to the tag information.
For both $J_{ic}$ and $J_{ci}$, we choose 0.2 as the margin parameter $\alpha$.
Finally, all networks are trained from scratch using SGD with weight decay of $1e^{-4}$, momentum of 0.9 and batch size of 512 on 8 V100 GPUs.
We train image encoder and text encoder with an initial learning rate of 0.03 and $4e^{-5}$, and adopt the cosine learning schedule. 
Training epochs are 200 and 60 for COCO and Stock, respectively.

\noindent\textbf{Downstream Tasks.}
Since our focus is visual representation learning, we only evaluate the pre-trained ResNet-50 backbone whose weights are denoted as $\theta_{q}$ while detaching all other modules in our framework.
In particular, 
to evaluate the COCO pretrained visual backbone, we perform ImageNet~\cite{DBLP:conf/cvpr/DengDSLL009} classification, PASCAL VOC~\cite{DBLP:journals/ijcv/EveringhamEGWWZ15} object detection, COCO instance segmentation tasks under the same setting and transfer learning protocols as in~\cite{DBLP:conf/cvpr/He0WXG20, DBLP:journals/corr/abs-2006-06666}. 
Similarly, we also evaluate the Stock pretrained model on multiple downstream tasks including image tagging and cross-modal retrieval on Stock's test set to demonstrate effectiveness of our method in a large-scale dataset setting. %on large-scale datasets.

\subsection{Evaluation of the COCO pretrained model}
\noindent\textbf{Linear Classification on ImageNet.}
We evaluate the learned visual representations (generated from our model pretrained on the COCO dataset) on ImageNet-1K (IN-1K) classification task. 
We choose ResNet-50 architecture as our visual backbone for all competing methods.
The competing methods include three types: (1) supervised classification-based, (2) self-supervised, and (3) text-supervised. For supervised methods, we report the performance of competing methods trained on both full-sized IN-1K (1.28M images) and IN-100 (100K images) under label supervision. 
We construct IN-100 by randomly sample 100 images per class, which has the similar amount of images with COCO train set (118K). For self-supervised methods, we compare with MoCo and MoCo-v2, both pre-trained on COCO dataset with unlabeled image data. For text-supervised methods, we adopt recently proposed Virtex and $\text{ICMLM}_{tfm}$, and compare to the best performance from their original papers. 

We also evaluate another variant of our method -- with additional tag supervision. %pretraining with tag supervision. 
Following the same transfer learning protocols %consistent with
as the competing methods, we train our linear classifier on features extracted from our frozen visual backbone on ImageNet-1K (IN-1K).
In particular, we perform linear classification with a fully connected layer plus softmax on 2048-d global average pooled features extracted from the last layer of the COCO pre-trained visual backbone. 
We train on the IN-1K training set and report top-1 validation accuracy on the validation split, without performing any specific data augmentation.
We train with batch size of 256 on 8 V100 GPUs for 60 epochs. 
We use SGD with momentum 0.9, weight decay 0 and learning rate 30, which is divided by 5 at 30th, 40th, and 50th epoch.

As shown in Table~\ref{tab:imagenet:result}, our pretrained visual backbone outperforms self-supervised methods, which demonstrates that we effectively enhance the visual feature quality by utilizing the semantic information from other modalities.
Our method trained with captions outperforms both VirTex and ICMLM by 2.1 \textit{p.p} and 3.0 \textit{p.p} respectively.  Note that ICMLM uses heavier data augmentations in the linear classifier training stage and VirTex uses a longer training schedule in its pre-training stage than ours, which might lead to a performance gain in the IN-1K classification task. 
However, our method still achieves better performance, which suggests that it benefits from our design for implicit similarity preservation within individual modalities.
We also train the model from scratch without using pre-trained BERT model.
The scratch model yields slight lower performance than the default one and still consistently outperforms VirTex.
In addition to comparisons with the state-of-the-art, we also observe that we can further improve the performance by 0.4 \textit{p.p} by %involving
leveraging COCO's tag information.
More importantly, when comparing with supervised methods %using
that use similar amounts of pre-training data in IN-100, our method still performs better, even though 
the IN-1K classification task may unfairly favor supervised methods~\cite{DBLP:journals/corr/abs-2006-06666}.

\begin{table}[tb]
\begin{center}
\footnotesize
\caption{Fully-, Un- and Text-supervised methods trained with ResNet50 backbones. 
We report top-1 obtained by linear classifier (on IN-1K) using pre-extracted features.}
\label{tab:imagenet:result}
\begin{tabular}{ccccc}
%{lccccccccc}
\toprule
&\multicolumn{1}{c}{{Model}}
&\multicolumn{1}{c}{{Pretrain Dataset}}
&\multicolumn{1}{c}{Supervision}
&\multicolumn{1}{c}{Top-1(\%)}\\
\midrule
&IN-Sup &IN-1K &Label &76.1  \\
&IN-Sup &IN-100 &Label &53.3 \\
\midrule
&MoCo~\cite{DBLP:conf/cvpr/He0WXG20} &COCO &NA  &44.5 \\
&MoCov2~\cite{DBLP:journals/corr/abs-2003-04297} &COCO &NA &49.3 \\
\midrule
&VirTex~\cite{DBLP:journals/corr/abs-2006-06666} &COCO &Caption &52.8 \\
&$\text{ICMLM}_{tfm}$~\cite{DBLP:conf/eccv/SariyildizPL20} &COCO &Caption &51.9 \\
&Ours &COCO &Caption &54.9 \\
&Ours(scratch) &COCO &Caption &54.6 \\
&Ours(with tag) &COCO &Caption+Tag &55.3 \\
\bottomrule
\end{tabular}
\end{center}
\vspace{-2em}
\end{table}
\begin{table}[tb]
\begin{center}
\footnotesize
\caption{Object Detection on PASCAL VOC.}
%We train Faster R-CNN detectors with ResNet-50-C4 backbones.}
\label{tab:pascal:result}
\begin{tabular}{cccccc}
%{lccccccccc}
\toprule
&\multicolumn{1}{c}{{Model}}
&\multicolumn{1}{c}{{Pretrain Dataset/Epochs}}
&\multicolumn{1}{c}{$\text{AP}_{50}$}
&\multicolumn{1}{c}{$\text{AP}$}
&\multicolumn{1}{c}{$\text{AP}_{75}$} \\
\midrule
&Random Init &NA/NA &60.2 &33.8 &33.1\\
&In-Sup &IN-1K~/~90 &81.6 &54.3 &59.7\\
\midrule
&MoCo~\cite{DBLP:conf/cvpr/He0WXG20} &IN-1K~/~200 &81.5 &55.9 &62.6\\
&MoCo-v2~\cite{DBLP:journals/corr/abs-2003-04297} &IN-1K~/~200 &82.4 &57.0 &63.6 \\
&MoCo~\cite{DBLP:conf/cvpr/He0WXG20} &COCO~/~200 &75.4 &47.5 &51.1\\
&MoCo-v2~\cite{DBLP:journals/corr/abs-2003-04297} &COCO~/~200 &75.5 &48.4 &52.1\\
\midrule
&VirTex~\cite{DBLP:journals/corr/abs-2006-06666} &COCO~/~1000 &81.4 &55.6 &61.5\\
&VirTex*~\cite{DBLP:journals/corr/abs-2006-06666} &COCO~/~200 &80.2 &54.8 &60.9\\
&Ours &COCO~/~200 &80.8 &55.6 &61.9 \\
&Ours(scratch) &COCO~/~200 &80.7 &55.4 &61.5 \\
&Ours(scratch) &COCO~/~1000 &82.1 &56.1 &62.4 \\
&Ours(with tag) &COCO~/~200  &81.8 &55.8 &61.7 \\
\bottomrule
\end{tabular}
\end{center}
\vspace{-2em}
\end{table}

\begin{table*}[thb]
\begin{center}
\footnotesize
\caption{Instance Segmentation on COCO: We use Mask R-CNN with ResNet-50-FPN backbones.}
\label{tab:cocoseg:result}
\begin{tabular}{ccccccccc}
%{lccccccccc}
\toprule
&\multicolumn{1}{c}{{Model}}
&\multicolumn{1}{c}{{Pretrain Dataset~/~Epochs}}
&\multicolumn{1}{c}{Box $\text{AP}$}
&\multicolumn{1}{c}{Box $\text{AP}_{50}$}
&\multicolumn{1}{c}{Box $\text{AP}_{75}$} 
&\multicolumn{1}{c}{Mask $\text{AP}$}
&\multicolumn{1}{c}{Mask $\text{AP}_{50}$}
&\multicolumn{1}{c}{Mask $\text{AP}_{75}$} \\
\midrule
&Random Init &NA~/~NA &36.7 &56.7 &40.0 &33.7 &53.8 &35.9\\
&In-Sup &IN-1K~/~90 &41.1 &62.0 &44.9 &37.2 &59.1 &40.0\\
\midrule
&MoCo~\cite{DBLP:conf/cvpr/He0WXG20} &IN-1K~/~200 &40.8 &61.6 &44.7 &36.9 &58.4 &39.7\\
&MoCo-v2~\cite{DBLP:journals/corr/abs-2003-04297} &IN-1K~/~200 &41.5 &62.2 &45.7 &37.4 &59.6 &40.5 \\
&MoCo~\cite{DBLP:conf/cvpr/He0WXG20} &COCO~/~200 &38.5 &58.5 &42.0 &35.0 &55.6 &37.5\\
&MoCo-v2~\cite{DBLP:journals/corr/abs-2003-04297} &COCO~/~200 &39.8 &59.6 &43.1 &35.8 &56.9 &38.8\\
\midrule
&VirTex~\cite{DBLP:journals/corr/abs-2006-06666} &COCO~/~1000 &40.9 &61.7 &44.8 &36.9 &58.4 &39.7\\
&VirTex*~\cite{DBLP:journals/corr/abs-2006-06666} &COCO~/~200 &39.6 &60.9 &44.0 &36.0 &57.6 &38.9\\
&Ours &COCO~/~200 &41.1 &61.8 &44.9 &36.9 &58.2 &40.0 \\
&Ours(with tag) &COCO~/~200 &41.2 &61.9 &44.9 &37.1 &58.9 &40.1 \\
\bottomrule
\end{tabular}
\end{center}
\vspace{-2em}
\end{table*}
\noindent\textbf{Object Detection on PASCAL VOC.}
In addition to evaluating features extracted from frozen visual backbones, we follow anther common protocol~\cite{DBLP:journals/corr/abs-2006-06666}: fine-tuning the entire visual backbone for the object detection task on PASCAL VOC. For the competing methods, we choose the \textit{Random Init} method as a simple baseline where the visual backbone is randomly initialized and trained on the downstream task. 
For IN-1K pretrained methods, we compare with both supervised and unsupervised (MoCo, MoCo-v2) methods. We also compare with MoCo and MoCo-v2 models trained on COCO with default hyperparameters. 
When comparing with VirTex, we report the performance from (1) the original paper (pre-trained with 1000 epochs), (2) the model pretrained using the official code \footnote{https://github.com/kdexd/virtex} and the default hyperparameters except that the epochs are set to 200.
We train Faster R-CNN~\cite{DBLP:conf/nips/RenHGS15} models with ResNet-50-C4 backbones on VOC trainval 07+12 split, while evaluating on test2007 split using Detectron2~\footnote{https://github.com/facebookresearch/detectron2}. 
In particular, during the fine-tuning stage of total 24K iterations, we warmup learning rate for first 100 iterations to 0.02, which is then decayed by 10 at 18K and 22K iterations.
The batch size is 16 and BN layers are synchronized (SyncBN)~\cite{DBLP:conf/cvpr/PengXLJZJYS18} across 8 V100 GPUs.

As shown in Table~\ref{tab:pascal:result}, our method significantly outperforms self-supervised methods which use COCO for pretraining. 
This suggests that useful semantic information is captured by our unique intra-modal pre-training scheme, which yields the visual backbone with better transferrability for the VOC object detection task.
When comparing to VirTex under the same pre-training epochs of 200, our method consistently performs better.
We also observe that we achieve comparable APs with VirTex (1000 epochs) while using 5$\times$ fewer pre-training epochs. 
When both trained for 1000 epochs from scratch, our method still consistently outperforms VirTex. 
These results suggest we benefit from the intra-modal similarity preservation that is not included in Virtex's design, which merely relies on a cross-modal proxy task during pre-training.

\noindent\textbf{Instance Segmentation on COCO.}
We evaluate our method on the instance segmentation task of COCO, while choosing the same protocol and competing methods for VOC object detection.
Following the setting %with
used by VirTex, we train Mask R-CNN~\cite{DBLP:conf/iccv/HeGDG17} models with ResNet-50-FPN~\cite{DBLP:conf/cvpr/LinDGHHB17} backbones on train2017split and evaluate on val2017 split. 
We adopt the 2$\times$ schedule of total 180K iterations with the initial learning rate of 0.02, which is decayed by 10 at 120K and 160K iterations.
The batch size is 16 across 8 V100 GPUs and SyncBN is used.
As shown in Table~\ref{tab:cocoseg:result}, our method performs slightly better than VirTex.
Both our model and VirTex outperform the self-supervised methods which are also trained with COCO.
Both methods achieve comparable results with the  methods that are pre-trained on IN-1K with 10$\times$ more data.
%We conclude that both ours and Virtex benefit from the additional information, leading to significant data efficiency.

\noindent\textbf{Cross-modal Retrieval on COCO.}
We evaluate the learned visual representation on both image-to-text and text-to-image retrieval tasks on COCO.
For a fair comparison with other methods which are pre-trained without textual encoder, we adopt 1-layer GRU which is randomly initialized as the sentence encoder~\cite{DBLP:conf/bmvc/FaghriFKF18} for all methods. 
%As such, we only evaluate the pre-trained weights of visual backbones.
This enables us to focus on comparing only the visual backbones.
All visual backbones are ResNet-50 and generate 2048-d global pooled features, which are then mapped to 1024-d by fully connected layers. For the GRU textural encoder, we set the word embedding size to 300 and the dimensionality of the embedding space to 1024.
We train the visual and textual encoders using the VSE++~\cite{DBLP:conf/bmvc/FaghriFKF18} loss on 113K images with 5 captions each and evaluate on 1K images.
We use the Adam optimizer~\cite{DBLP:journals/corr/KingmaB14} and set the batch size as 128. 
We follow the transfer protocol as in~\cite{DBLP:conf/bmvc/FaghriFKF18}: train with a fixed image encoder with learning rate 2$e^{-4}$ for 15 epochs, and then decrease the learning rate to 2$e^{-5}$ for %another
the next 15 epochs.
As for evaluation, we use the same evaluation protocols as in~\cite{DBLP:conf/bmvc/FaghriFKF18}: (1) $R@K$, defined as the percentage of queries in which the corresponding image is contained in the first $K$ retrieved results. The higher this value, the better. (2) Med r, which is the median rank of the first retrieved ground-truth sentence or image. The lower its value, the better.
The results are in Table~\ref{tab:cocosearch}. We see that our method consistently performs better than all competing methods. This suggests that the learned visual representation not only generalizes better on image-based downstream tasks but also on the cross-modal task, demonstrating the effectiveness of the implicit cross-modal similarity preservation objective.
\begin{table}[tb]
\begin{center}
\footnotesize
\caption{Cross-modal search on COCO 1K test-set.}
%: We train the ResNet-50 visual backbones and 1-layer GRU using VSE++ loss~\cite{DBLP:conf/bmvc/FaghriFKF18}. IN-sup and MoCo-v2 are pre-trained using IN-1K while VirTex and ours are pre-trained on COCO.}
\label{tab:cocosearch}
% \begin{table}[!h]
% \label{T:equipos}
% \begin{center}
\begin{tabular}{ccccccc}
\toprule
\textbf{Method} & \multicolumn{3}{c|}{\textbf{Image-to-Text}}  & \multicolumn{3}{ c}{\textbf{Text-to-Image}} \\ 
\cline{2-7}
& \textbf{R@1} & \textbf{R@10} & \textbf{Med r} & \textbf{R@1} & \textbf{R@10} & \textbf{Med r} \\
\hline
IN-sup &57.9  &92.7  &1.0 &42.8 &87.0 &2.0   \\
\midrule
MoCo-v2~\cite{DBLP:journals/corr/abs-2003-04297} &51.6 &90.0  &1.0 &39.0 &84.8 &2.0     \\
\midrule
VirTex~\cite{DBLP:journals/corr/abs-2006-06666} &58.1  &93.0  &1.0 &44.0 &88.5 &2.0     \\
Ours &58.4  &93.4  &1.0 &45.1 &90.0 &2.0   \\
\bottomrule
\end{tabular}
\end{center}
\vspace{-2em}
\end{table}
\subsection{Evaluation on the Stock dataset}
We evaluate our method on the Stock image dataset and evaluate on its image tagging and cross-modal retrieval task.
\begin{table}[tb]
\begin{center}
\footnotesize
\caption{Image tagging on Stock 50K test-set.}
\label{tab:stocktag}
% \begin{table}[!h]
% \label{T:equipos}
% \begin{center}
\begin{tabular}{ccccc}
\toprule
%\textbf{Method} & \multicolumn{3}{c}{\textbf{Image-to-Text}}  & \multicolumn{3}{ c}{\textbf{Text-to-Image}} \\ 
%\cline{2-7}
\textbf{Method} & \textbf{mIOU@5} & \textbf{mIOU@10} & \textbf{mIOU@15} & \textbf{mIOU@20} \\
\hline
Stock-sup &12.86\%  &13.93\%  &14.88\% &15.74\%    \\
Ours &13.81\%  &14.69\%  &15.55\% & 16.42\%     \\
\bottomrule
\end{tabular}
\end{center}
\vspace{-1em}
\end{table}
\noindent\textbf{Image Tagging on Stock.}
Similar with image classification task, we train on features extracted from the frozen visual backbone. 
In particular, we map the tags to 4096-d feature using Pointwise Mutual Information (PMI) embedding~\cite{DBLP:journals/coling/ChurchH90}. 
We use a 2-layer MLP (hidden layer is 2048-d with ReLU) to map image backbone feature to 4096-d.
We train the MLP using the cosine similarity loss between pre-extracted tagging features and image features in the common embedding space. We compare with the supervised method, denoted as Stock-sup, which directly trains the backbone on the tagging task. The evaluation metric is mIOU$@K$, which is measured by the average overlaps between top-K predicted tags (pred) and ground-truth tags (gt) over all test samples N, \textit{i.e.} $\sum_{i=1}^{N}\frac{|pred \bigcap gt|}{|pred \bigcup gt|}/N$.
%Image tagging on Stock is an extremely  unfair  comparison  for our methods when comparing with Stock-sup which is trained explicitly for the downstream task
The Stock-sup model is trained directly for the tagging task and has an extremely unfair advantage, compared to the model trained with our method. However, as shown in Table~\ref{tab:stocktag}, our method consistently outperforms the Stock-sup model.

\noindent\textbf{Cross-modal Retrieval on Stock:}
We evaluate our method on the Stock cross-modal retrieval task, following the protocol in COCO cross-modal retrieval, except that we use 10K images for testing. Note that the competing method Stock-sup pre-trains the backbone on the Stock tagging task. As shown in Table~\ref{tab:stocksearch}, our method consistently outperforms the supervised baseline by a large margin.

\begin{table}[tb]
\vspace{-1em}
\begin{center}
\footnotesize
\caption{Cross-modal search on Stock 10K test-set.}
%We train the ResNet-50 visual backbones and 1-layer GRU using VSE++ loss.}
\label{tab:stocksearch}
% \begin{table}[!h]
% \label{T:equipos}
% \begin{center}
\begin{tabular}{ccccccc}
\toprule
\textbf{Method} & \multicolumn{3}{c|}{\textbf{Image-to-Text}}  & \multicolumn{3}{ c}{\textbf{Text-to-Image}} \\ 
\cline{2-7}
& \textbf{R@1} & \textbf{R@10} & \textbf{Med r} & \textbf{R@1} & \textbf{R@10} & \textbf{Med r} \\
\hline
Stock-sup &33.76  &71.22  &3 &30.64  &68.18  &4   \\
Ours &36.98  &74.02  &2 &33.83  &70.76  &3   \\
\bottomrule
\end{tabular}
\end{center}
\vspace{-2em}
\end{table}

\subsection{Ablation Study}
\noindent\textbf{Investigation on Separate MLP Design.}
Using distinct MLPs for intra- and inter-modal feature embedding with different dimensions is an essential design in our method.
We show the effectiveness of this design by varying the MLP head in our method (1) unified MLPs: we use shared MLP layers with output dimension of $\{128, 1024, 1152\}$. (2) distinct MLPs but with same dimensions of $\{128, 1024, 1152\}$. We train both baselines on COCO images and captions, and evaluate on the IN-1K image classification.
For baseline (1), we achieve $49.6\%$, $50.2\%$ and $52.3\%$ top-1 accuracy, respectively.
For baseline (2), the performance is $53.6\%$, $53.8\%$ and $53.9\%$.
Separate design consistently yields better visual features than unified design for IN-1K.
Our final design (128-d for intra-modal; 1024-d for inter-modal) performs best ($54.9\%$), which suggests our unified framework benefits from using a larger-sized common space for cross-modal correlation modeling and a relatively small one for self-supervision.

\noindent\textbf{Investigation on Tag Supervision.}
We show the effectiveness of the tag supervision term in visual learning by pre-training on COCO unlabeled images using (1) $J_{ii}$; (2) $J_{ii}$+$J_{tag}$. and evaluating on IN-1K image classification. Note that training with merely the $J_{ii}$ term is equivalent to MoCo-v2, which achieves 49.3$\%$ top-1 accuracy as shown in Table~\ref{tab:imagenet:result}. 
Training with the additional tag supervision term $J_{tag}$ further improve the accuracy to 50.2$\%$. 
The final model trained with all five losses achieves the best performance of 55.3$\%$, benefiting from the inter-modal training %with
that leverages both image and caption information. We also use the ICMLM-style POS tagging + noun filtering to form image labels, in which ImageNet classification accuracy drops from 55.3\% to 54.5 \% by ignoring tags.

\begin{table}[tb]
\begin{center}
\footnotesize
\caption{Cross-modal search on COCO without fine-tuning.}
\label{tab:cocosearch:nofine}
% \begin{table}[!h]
% \label{T:equipos}
% \begin{center}
\begin{tabular}{ccccccc}
\toprule
\textbf{Method} & \multicolumn{3}{c|}{\textbf{Image-to-Text}}  & \multicolumn{3}{ c}{\textbf{Text-to-Image}} \\ 
\cline{2-7}
& \textbf{R@1} & \textbf{R@10} & \textbf{Med r} & \textbf{R@1} & \textbf{R@10} & \textbf{Med r} \\
\hline
RS &0.1  &1.0  &650 &0.1 &1.0 &500   \\
Inter-modal &13.2 &42.2 &10.0 &9.5 &36.7 &16.0     \\
Multimodal &24.1  &66.0  &5.0 &18.6 &58.4 &7.0     \\
\bottomrule
\end{tabular}
\end{center}
\vspace{-2em}
\end{table}

\noindent\textbf{Investigation on Learned Textual Features.}
Even though the main focus of this paper is visual representations, we still generate useful textual features as a by-product.
We show this %using
via a COCO cross-modal retrieval downstream task.
We design three variants for comparison: 
(1) RS: we randomly pair the image and caption across 1K test data.
(2) Inter-modal: we train image and textual encoders merely with $J_{ic}$ plus $J_{ci}$ on images and captions. MLPs output dimensions are all set as 1024.
(3) Multimodal: we use the multimodal pre-trained visual/textual encoder for feature extraction. 
As shown in Table~\ref{tab:cocosearch:nofine}, we observe that even without fine-tuning, variant (2) and (3) still perform much better than random baseline (1), which suggests some useful textual information %is already
has already been captured by the model during the pre-training stage.

\section{Conclusion}
We propose a simple yet effective method to learn visual representations in a unified multimodal training framework, composed of two intra-modal and inter-modal learning paths with carefully designed contrastive losses.
Extensive experiments across various datasets and tasks demonstrate that we learn high-quality visual features with better scalability and transferability.
Since our framework is symmetric for all modalities (e.g. image and caption explored here), it has the flexibility to be extended to other modalities such as video and audio.

{\small
\bibliographystyle{ieee_fullname}
\bibliography{cvpr}
}

\end{document}